\documentclass[]{spie} 
\usepackage{cite}
\usepackage{amsmath,amssymb,amsfonts}
\usepackage{algorithmic}
\usepackage{graphicx}
\usepackage{textcomp}

\usepackage{multirow}
\usepackage{url}
\usepackage{hyperref}
\hypersetup{
    colorlinks=true,
    linkcolor=blue,
    citecolor=blue,
    urlcolor=blue,
    }
\usepackage{makecell}
\usepackage{placeins}

\begin{document}

\title{Morphology-Enhanced CAM-Guided SAM for Weakly Supervised Breast Lesion Segmentation}

\author{Xin Yue, Xiaoling Liu, Qing Zhao, Jianqiang Li, Changwei Song, Suqin Liu, Zhikai Yang, Guanghui Fu\textsuperscript{*}
    %\thanks{This study is supported by a research project from the Beijing Postdoctoral Research Foundation (No. 2022zz075). Guanghui Fu is supported by a Chinese Government Scholarship provided by the China Scholarship Council (CSC).}
    %\thanks{Xin Yue, Xiaoling Liu, Qing Zhao, Jianqiang Li, Changwei Song and Suqin Liu are with School of Software Engineering, Beijing University of Technology, Beijing, China.}
    %\thanks{Zhikai Yang is with Department of Biomedical Engineering and Health, KTH Royal Institute of Technology, Stockholm, Sweden}
    %\thanks{Guanghui Fu is with Sorbonne Universit\'{e}, Institut du Cerveau – Paris Brain Institute - ICM, CNRS, Inria, Inserm, AP-HP, H\^{o}pital de la Piti\'{e}-Salp\^{e}tri\`{e}re, Paris, France (e-mail: guanghui.fu@inria.fr).}
    %\thanks{Corresponding author: Guanghui Fu (\url{guanghui.fu@inria.fr})}
}

\authorinfo{
Xin Yue, Xiaoling Liu, Qing Zhao, Jianqiang Li, Changwei Song and Suqin Liu are with School of Software Engineering, Beijing University of Technology, Beijing, China.
Zhikai Yang is with Department of Biomedical Engineering and Health, KTH Royal Institute of Technology, Stockholm, Sweden
Guanghui Fu is with Sorbonne Universit\'{e}, Institut du Cerveau – Paris Brain Institute - ICM, CNRS, Inria, Inserm, AP-HP, H\^{o}pital de la Piti\'{e}-Salp\^{e}tri\`{e}re, Paris, France (e-mail: guanghui.fu@inria.fr).
Corresponding author: Guanghui Fu (\url{guanghui.fu@inria.fr})}

\maketitle

\begin{abstract}
Ultrasound imaging plays a critical role in the early detection of breast cancer. Accurate identification and segmentation of lesions are essential steps in clinical practice, requiring methods to assist physicians in lesion segmentation. However, ultrasound lesion segmentation models based on supervised learning require extensive manual labeling, which is both time-consuming and labor-intensive. In this study, we present a novel framework for weakly supervised lesion segmentation in early breast ultrasound images. Our method uses morphological enhancement and class activation map (CAM)-guided localization. Finally, we employ the Segment Anything Model (SAM), a computer vision foundation model, for detailed segmentation. This approach does not require pixel-level annotation, thereby reducing the cost of data annotation. The performance of our method is comparable to supervised learning methods that require manual annotations, achieving a Dice score of 74.39\% and outperforming comparative supervised models in terms of Hausdorff distance in the BUSI dataset. These results demonstrate that our framework effectively integrates weakly supervised learning with SAM, providing a promising solution for breast cancer image analysis. The code for this study is available at: \url{https://github.com/YueXin18/MorSeg-CAM-SAM}.
\end{abstract}

%\begin{IEEEkeywords}
%\end{IEEEkeywords}

\section{Introduction}
\label{sec:intro}

%\IEEEPARstart{B}{reast} cancer is one of the most common malignant tumors affecting women, with its incidence rising annually~\cite{sun2017risk}. Globally, in 2020, there were approximately 2.26 million cases of breast cancer, leading to around 685 thousand deaths~\cite{who2023breast}. This makes it the leading cause of cancer-related mortality among women worldwide~\cite{wilkinson2022understanding}. The World Health Organization launched the Global Breast Cancer Initiative in 2021, aiming to tackle this significant health challenge~\cite{wilkinson2022understanding}.
Breast cancer is one of the most common malignant tumors affecting women, with its incidence rising annually~\cite{sun2017risk}. Globally, in 2020, there were approximately 2.26 million cases of breast cancer, leading to around 685 thousand deaths~\cite{who2023breast}. This makes it the leading cause of cancer-related mortality among women worldwide~\cite{wilkinson2022understanding}. The World Health Organization launched the Global Breast Cancer Initiative in 2021, aiming to tackle this significant health challenge~\cite{wilkinson2022understanding}.
Globally, three main inequities affect breast cancer care~\cite{who2023initiative}: late diagnosis, often at advanced stages; inadequate services, including limited diagnostic and treatment facilities; and low coverage, particularly in the inclusion of breast cancer in Universal Health Coverage (UHC).
The risk factors for breast cancer are multifaceted. Age is a primary risk factor, with older women exhibiting the highest incidence rates~\cite{benson2009early}. Furthermore, genetic factors play a role in 5-10\% of cases, notably mutations in the BRCA1 or BRCA2 genes~\cite{wilkinson2022understanding}. Early detection and prompt treatment are vital, as the five-year survival rate can exceed 90\% with early diagnosis~\cite{islami2018proportion}. In terms of diagnostics, while manual examinations are common, three key imaging techniques are employed: mammography, magnetic resonance imaging (MRI), and ultrasound. Ultrasound is particularly effective for examining dense breasts, as it provides detailed insights into the morphology, orientation, internal structure, and margins of lesions~\cite{jafari2018breast, kuhl2017supplemental}. These assessments are crucial for distinguishing between benign and malignant breast lesions~\cite{iranmakani2020review}. Ultrasound stands out as a highly sensitive, non-invasive, radiation-free, and cost-effective method for early breast cancer detection and diagnosis, especially in dense breast tissue~\cite{guo2018ultrasound}.

Computer-aided diagnosis (CAD) has emerged as a key research priority for radiologists, particularly for enhancing the efficiency of interpreting ultrasound images~\cite{dromain2013computed}. CAD systems can autonomously analyze lesion characteristics and differentiate them from normal tissues. However, the automatic detection of breast tumors presents challenges, notably due to their irregular shapes and blurred boundaries. Current research in breast ultrasound (BUS) image lesion segmentation falls into two main categories: traditional methods that rely on predefined features~\cite{zahoor2020breast} and those based on deep learning~\cite{liu2019deep, van2019deep}. Traditional segmentation approaches, such as region growing-based~\cite{kolahdoozi2017fuzzy, punitha2018benign, shrivastava2017automated, el2018breast}, threshold-based~\cite{singh2018breast, zebari2020improved, yang2023multi, suradi2021breast}, and clustering-based methods~\cite{muhammad2020region, huang2017breast}, effectively capture contour information of lesions. However, they often struggle with generalization, especially in cases of lesions with fuzzy and irregular boundaries. In contrast, deep learning-based methods have shown significant progress in detecting breast lesions. A notable development in this area is the introduction of an end-to-end convolutional neural network (CNN), UNet, specifically designed for medical image segmentation~\cite{ronneberger2015u}. Following its introduction, several neural network architectures similar to UNet, such as~\cite{vidal2022u, byra2020breast, chen2022aau, ning2021smu, sun2020aunet}, have been developed, demonstrating enhanced abilities in segmenting breast lesions.

Most deep learning methods in medical imaging operate under fully supervised scenarios, where their performance heavily depends on the quantity of pixel-level labels~\cite{zhou2018brief}. However, annotating medical images requires specialized medical knowledge, including disease diagnosis and understanding of anatomical structures, making accurate and comprehensive annotation challenging. Consequently, achieving better segmentation results with less costly image annotation has become a focal point in medical image segmentation research. Researchers are increasingly exploring methods that can efficiently utilize imperfect data or labels, such as weakly supervised segmentation algorithms~\cite{tajbakhsh2020embracing, zhou2018brief}. Some approaches use classification networks and class activation maps (CAM)~\cite{zhou2016learning}, which identify semantic features and help initially localize lesions~\cite{li2022deep, altini2022ndg}. These methods not only facilitate lesion detection but also aid in understanding the classification model’s predictions~\cite{groen2022systematic}, allowing researchers and users to examine the model's decision-making basis~\cite{fu2021attention, wang2022diagnosis}. However, CAMs typically provide only a rough estimate of the predicted areas, and their ability to precisely detect lesion boundaries, especially edges, is limited~\cite{jiang2021layercam, bae2020rethinking}. 
Furthermore, the effectiveness of weakly supervised methods largely depends on the accuracy of pseudo-labels. These output generated by CAM, can be influenced by surrounding background noise~\cite{sun2021ecs}.
The recently introduced Segment Anything Model (SAM)~\cite{kirillov2023segment} represents an advancement in addressing these challenges. Trained on over 11 million images with 1 billion masks (SA-1B), SAM is capable of zero-shot segmentation on unseen images using various prompts such as bounding boxes, points, and text~\cite{chen2023segment}. However, its application to medical images, which often feature complex biological tissues with diverse shapes and characteristics, differs from natural images. Therefore, direct segmentation of lesions using SAM in medical contexts has not yet yielded optimal results~\cite{mazurowski2023segment, shi2023generalist}. 
More critically, the SAM requires prompt-based input to perform medical image segmentation tasks, making it difficult to automate and challenging for scalable applications in clinical settings.
% This highlights the need for continued research and adaptation of these advanced segmentation tools for specific challenges in medical imaging.

To overcome the limitations of current methods, we propose a novel weakly supervised lesion segmentation framework comprising four main modules in breast ultrasound imaging: a traditional segmentation module based on morphology, a semantic information extraction and lesion localization module, an information fusion module, and a SAM fine-grained segmentation module. The traditional segmentation module utilizes morphology to perform initial segmentation and extract contour information from medical images, focusing on the shape, edge, and direction of lesions. The semantic information extraction and lesion localization module, leveraging image-level category labels, trains a classification network and achieves a fuzzy localization of lesions through the heat map provided by CAM. The information fusion module then adeptly combines the outputs from these two modules, generating a more comprehensive lesion area. Finally, SAM utilizes this area as a prompt for segmenting lesions, refining the segmentation process and enhancing the results through post-processing. This integrated approach aims to address the challenges in weakly supervised breast lesion segmentation by combining traditional and advanced techniques for more accurate and efficient results. The key contributions of this paper are outlined as follows: 
\begin{itemize}
  \item This paper introduces a novel integration of SAM with weakly supervised methods for breast lesion segmentation. It can refine segmentation regions when fed with regions derived from CAM or similar techniques. This ability is particularly beneficial in scenarios with limited training data, ensuring improved segmentation outcomes.
  \item The proposed segmentation framework integrates prior knowledge of lesion morphology, semantic features from medical images, and the precise segmentation capabilities of SAM. By merging these diverse methodologies, the framework is able to learn more comprehensive lesion features, leveraging the combined strengths of each approach.
  \item The framework has been evaluated on two publicly available datasets, showcasing comparable performance as fully supervised learning approaches, as indicated by the overlap of the confidence intervals. And it outperformed other weakly supervised segmentation method. 
\end{itemize}

\section{Related work}
\label{sec:related}

\subsection{Weakly supervised segmentation methods}
Deep learning, particularly through architectures like UNet~\cite{ronneberger2015u} and DeepLabv3+, has revolutionized lesion segmentation in medical imaging. UNet, known for its encoder-decoder structure and skip connections, simplifies the segmentation process by eliminating complex manual feature extraction. DeepLabv3+ extends these capabilities with dilated convolution, effectively handling variations in lesion size and shape~\cite{chen2018encoder}. This adaptability makes it especially suitable for complex medical image segmentation tasks.

However, supervised learning based lesion segmentation relies on high-quality, pixel-level annotated datasets, which makes this field challenging. To achieve high-quality segmentation with easier and cost-effective annotations, researchers are exploring weakly supervised strategies, such as interaction-based methods. These methods involve user participation in selecting regions, marking boundaries, and refining labels, guiding the algorithm for better segmentation. For instance, Roth et al.~\cite{roth2021going} used a random walk algorithm with user clicks to train a convolutional network, enhancing segmentation with custom loss functions and attention mechanisms. Pinheiro and Collobert~\cite{pinheiro2015image} proposed a method using pixel-level labels and back-propagation of errors for more accurate weakly supervised segmentation. These approaches offer promising alternatives to resource-heavy fully supervised methods, especially when high-quality annotated datasets are scarce.

Weakly supervised semantic segmentation has gained significant attention in recent times, especially with the advancement of class activation mapping (CAM) techniques~\cite{zhou2016learning}. Various researchers have proposed innovative methods to enhance the accuracy and overcome the inherent limitations of CAM-based approaches. For instance, Chen et al.~\cite{chen2022c} introduced the causal CAM (C-CAM) method, addressing the challenge of unclear object boundaries between foreground and background. C-CAM operates on two causal chains: the category-causal chain, which relates to how image content affects categories, and the anatomical-causal chain, focusing on anatomical structures influencing organ segmentation. This method has been thoroughly tested across three public medical image datasets. Zhong et al.~\cite{zhong2021dap} proposed a combination of CAM and weakly supervised detection-aware pre-training (DAP). This approach leverages weakly labeled categorical data for pre-training and transforms the categorical dataset into a detection dataset via a weakly supervised target localization method based on class-activation mapping. This enables the pre-trained models to be location-aware and capable of predicting bounding boxes. Ahn and Kwak~\cite{ahn2018learning} proposed AffinityNet which generates accurate segmentation labels for training images based solely on image-level class labels. It combines semantic features with random walking to modify CAM and produce segmentation labels. These methods exemplify the innovative approaches in weakly supervised learning for medical image lesion detection and segmentation, significantly reducing the cost and subjectivity associated with manual labeling. However, weakly supervised methods primarily rely on image-level labeling, which can lead to the model learning imprecise features, impacting the final segmentation's accuracy. Additionally, these methods might not fully exploit all available information in complex images, particularly with intricate lesion structures or boundary cases. To address these challenges, the integration of boundary-refined gain tools is crucial. It can be a balanced and effective strategy for dealing with the complexities and nuances of medical image segmentation.

\subsection{SAM-based segmentation methods}
The Segment Anything Model (SAM)~\cite{kirillov2023segment} has recently gained considerable attention in the computer vision community for its remarkable zero-shot image segmentation capabilities. SAM, a model capable of generalizing to unfamiliar objects and images without additional training, incorporates the prompt paradigm from natural language processing into computer vision. This approach enables accurate image segmentation based on input prompts, such as points or boxes, and can generate masks for all objects in an image. However, SAM is primarily optimized for natural images, and its direct application in medical image segmentation has proven less accurate, posing significant challenges in this domain~\cite{mazurowski2023segment, ma2023segment}. Addressing these issues, researchers are focusing on adapting SAM for medical imaging. Ma and Wang et al.~\cite{ma2023segment} developed MedSAM, trained on a large-scale dataset containing over 1 million medical image-mask pairs. MedSAM is adept at segmenting various anatomical structures and lesions across different medical imaging modalities, offering a balanced mix of automation and customization. Similarly, Wu et al.~\cite{wu2023medical} introduced the Medical SAM Adapter, which integrates medical-specific knowledge into SAM using parameter-efficient adaptive fine-tuning techniques, significantly enhancing the original model. Fine-tuning SAM to suit specific datasets enhances its adaptability and ability to capture relevant features, improving performance on both familiar and unseen data~\cite{shi2023generalist}. This process allows SAM to learn more generalizable features, aiding its performance across diverse samples and scenarios. However, fine-tuning may demand substantial computational resources, such as increased training time and storage, which could limit its applicability. Moreover, a fine-tuned SAM might not generalize well across various types of medical images or segmentation tasks due to domain-specific adaptations. Recent studies have shown that using prompt methods with SAM can markedly improve its performance. Chen et al.~\cite{chen2023rsprompter} introduced RSPrompter, a novel prompt learning technique, to guide SAM in generating semantic instance-level masks, particularly enhancing its capabilities in remote sensing image instance segmentation. Deng et al.~\cite{deng2023sam} proposed the SAM-U framework, incorporating multi-frame prompts to achieve more accurate medical image segmentation. Compared to fine-tuning, prompt methods can reduce the dependency on large quantities of accurate labels. By integrating prompts into training, the model can achieve commendable performance with a relatively limited number of labels. Additionally, prompt methods enhance the model's interpretability, offering a more comprehensible and user-friendly approach to image segmentation.

These research demonstrate the potential and effectiveness of SAM, a generalized segmentation model that can be extended to the medical image field.

\section{Methods} \label{sec:methods}
\begin{figure*}
  \centering
  \includegraphics[width=1\linewidth]{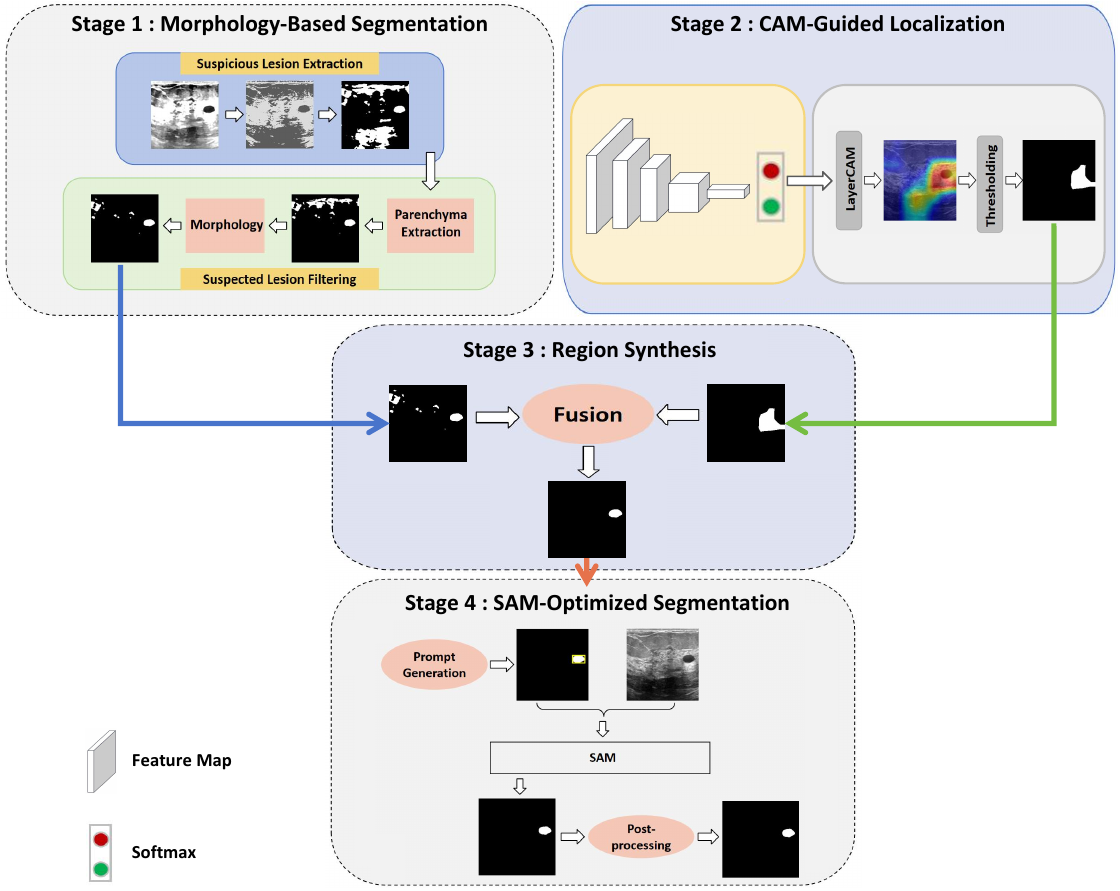}
  \caption{The proposed framework of our model consists of four key stages: in stage 1, we perform a preliminary image segmentation focusing on morphological features. This is followed by the generation of a heat map for lesion localization using a CAM-based classification network, as shown in stage 2. Subsequently, the features from the previous stages were synthesized as shown in stage 3 and generated as a bounding box prompt in stage 4 for detailed segmentation using the SAM. }
  \label{fig:model_architecture}
\end{figure*}

In this study, we proposed a weakly supervised lesion segmentation method for breast ultrasound images, inspired by the framework outlined in Liu et al.~\cite{liu2023anatomy}. The proposed method shown in Figure~\ref{fig:model_architecture}, commences with an initial segmentation of medical images, utilizing morphological knowledge like lesion shape and edges. Then followed by lesion localization obtained from an image classification model and CAM. Subsequently, we flexibly integrate the outcomes of traditional morphological segmentation with those of lesion localization. The process concludes with the application of the SAM and various post-processing techniques to refine the segmentation results to remove the topological error.

\subsection{Traditional segmentation based on morphological feature} \label{sec:methods:morphology}

Initially, our process begins with segmenting the medical image based on key characteristics of the lesions, such as shape, edges, and orientation. To achieve this, we employ the K-means algorithm to cluster the pre-processed image. Then, we apply thresholding to isolate all suspected lesions. These suspected lesions are then meticulously filtered using morphological knowledge to ensure precision in the segmentation.

\subsubsection{Suspicious lesion extraction}
The breast ultrasound images frequently exhibit low overall brightness, with grayscale values primarily confined to a lower range. To counteract this, we implement an automatic color enhancement (ACE) technique, as described by Getreuer~\cite{getreuer2012automatic}. This ACE algorithm enhances image contrast by assessing the brightness and interrelationships between a target pixel and its adjacent pixels. Within this context, let $I$ represent a specific color channel of an RGB breast ultrasound (BUS) image, defined over the domain $\Omega$. The intensities in this channel are normalized within the range $[0, 1]$. The ACE process is applied independently to each of the three color channels, facilitating chromatic aberration correction in the BUS image as shown in Equation~\ref{eq:ace:1}.
\begin{equation}
\begin{aligned}
R(x)=\sum_{y\in\Omega\backslash x}\frac{s_{\alpha}(I(x)-I(y))}{||x-y||}, x\in \Omega
\end{aligned}
\label{eq:ace:1}
\end{equation}
where $y\in\Omega\backslash x$ represents $\{\ y\in\Omega\ y\neq x\}\ $, $||x-y||$represents the Euclidean distance between $x$ and $y$. The slope function, $s_{\alpha}(*)$, plays a pivotal role in adapting to local image contrasts. It is designed to enhance minor variations and enrich significant ones, effectively compressing or expanding the dynamic range based on local image content. In the subsequent stage, we compute the enhancement channel by normalizing $R$ to the range $[0, 1]$ as shown in Equation~\ref{eq:ace:2}. This normalization is crucial for maintaining consistency in contrast enhancement across the image.
\begin{equation}
\begin{aligned}
L(x)=\frac{R(X)-\min R}{\max R-\min R}
\end{aligned}
\label{eq:ace:2}
\end{equation}
Following ACE, we employ the K-means clustering algorithm, an iterative method that groups the image data into $k$ distinct clusters based on a distance formula. The objective function $V$ depicted in Equation~\ref{eq:kmeans}, is used to achieve optimal clustering. Here, $k$ signifies the number of clusters, $n$ the number of points in each cluster, and $c_i$ the centroid of each cluster. The term $(x_{j}-c_{i})^{2}$ quantifies the distance between a data point $x_j$ and the centroid $c_i$ of its cluster.
\begin{equation}
\begin{aligned}
V=\sum_{i=1} ^{k}\sum_{j=1} ^{n}(x_{j}-c_{i})^{2}
\end{aligned}
\label{eq:kmeans}
\end{equation}
After clustering, we perform global threshold segmentation, dividing the image into target and background regions using global information. This step is essential in completing the extraction of all suspected lesions, setting the stage for subsequent detailed analysis and segmentation.

\subsubsection{Suspected lesion filtering}
Our filtering approach for suspected lesions in breast ultrasound (BUS) images involves a layered anatomical model, as advised by experienced radiologists. We classify BUS image structures into three layers: subcutaneous fat, breast parenchyma, and chest wall muscle. Early-stage breast cancer lesions, typically benign, are almost exclusively located in the parenchymal layer. Based on this a priori knowledge, we removed the suspected lesions located in the bottom 1/3 and top 1/10 of the image. 
%A radiologist with extensive clinical experience annotates this layer, assigning a mask $m={0, 1, 2}$ to each pixel, where $0$ denotes the subcutaneous fat layer, $1$ the breast parenchyma layer, and $2$ the chest wall muscle layer. This annotation facilitates the extraction of the breast parenchymal layer and the removal of suspected lesions from non-parenchymal areas.

For filtering suspected lesions within the parenchymal layer, we consider both the shape and aspect ratio of lesions, taking into account the textural characteristics of breast ultrasound images. Benign lesions like cystic nodules typically exhibit a hypoechoic, round or oval shape. We use morphological knowledge to filter out erroneously extracted tissues based on aspect ratio; benign lesions generally have a ratio between $0.2$ and $1$, while ducts and lobules have a lower ratio. We calculate the minimum enclosing rectangle for each lesion from the binary image obtained by K-means clustering and thresholding, determining height and width. Non-target areas are then filtered out using this morphological knowledge. The criterion for screening non-lesion areas, based on their aspect ratio, is defined as follows in Equation~\ref{eq:filter}:
\begin{equation}
\begin{aligned}
l_{i}=\begin{cases}
0, & if \quad ratio(l_{i})<0.2 \\
1, & otherwise \\
\end{cases}
\end{aligned}
\label{eq:filter}
\end{equation}
Here, $ratio\ (l_{i})$ denotes the aspect ratio of the $i$-th suspected lesion in the binary image. 

\subsection{CAM-Guided classification model for lesion localization}\label{sec:method:cam}
\subsubsection{Semantic information extraction}
To train a supervised learning segmentation model, the lesions in medical images require pixel-level annotation by professional medical experts. This process is time-consuming and costly, and the interpretation and annotation of medical images may be affected by the subjective judgment of experts. Different experts may give different annotations, which can lead to inconsistencies and uncertainty. Considering these existing problems, we use image-level labels to achieve semantic information extraction. 

In our study, we employ DenseNet for classifying breast ultrasound (BUS) images due to its ability to effectively utilize features from shallow layers with low complexity, aiding in achieving a smooth decision function with enhanced generalization performance. Specifically, we utilize the DenseNet-121 variant, which comprises 121 layers. The network's architecture begins with an initial convolutional layer designed for three input channels. This layer uses a $7 \times 7$ convolution kernel with a stride of 2 for extracting preliminary features. Subsequent processes include batch normalization and the application of ReLU activation functions. Spatial resolution is then reduced through a $3 \times 3$ maximum pooling operation. The neural network consists of four dense blocks and three transition layers. The dense blocks contain 6, 12, 24, and 16 convolutional blocks, respectively. Each block features tightly connected convolutional layers, incorporating both $1 \times 1$ and $3 \times 3$ convolutions. These layers process outputs from preceding layers, continuously integrating new features. Transition layers, positioned between the dense blocks, comprise a $1 \times 1$ convolution followed by a $2 \times 2$ average pooling layer to decrease the spatial dimension of the output. The culmination of dense blocks and transition layers leads to a batch normalization layer that normalizes the final feature set. This is followed by mapping these features to the number of output categories in the classification output layer. In this network, the initial convolutional layer extracts fundamental features like edges and textures. The subsequent convolutional layers within the dense blocks build upon these features, enhancing semantic information layer by layer. The dense connectivity ensures efficient feature reuse and information transmission, while the transition layers help maintain semantic information as they reduce the feature map size. Ultimately, global average pooling aggregates the feature maps into a comprehensive representation, capturing the overarching semantic information of the image.

Let $B=\{\ x_{i},y_{i}\}\ $ presents the BUS dataset, where $x_{i}$ represents the $i$-th image and $y_{i}$ is its corresponding image-level label, indicating the presence (or absence) of a lesion in image $x_{i}$. The data with lesions is represented as $P=((x_{i},y_{i})\in B|y_{i}=1)$, and the data without lesions is represented as $N=((x_{i},y_{i})\in B|y_{i}=0)$. 
The training process involves minimizing the binary cross entropy (BCE) loss, which is mathematically formulated as:
\begin{equation}
\begin{aligned}
L_{cls}=-\frac{1}{n}\sum_{i=1}^{n}y_{i}log\, s(f(x_{i}))+(1-y_{i})log(1-s(f(x_{i})))
\end{aligned}
\label{eq:loss}
\end{equation}
where $n$ is the number of samples, $f$ represents the classification network and $s(*)$ is the sigmoid function.

\subsubsection{Lesion localization}\label{sec:method:cam:lesion}
In weakly supervised breast lesion localization stage, we employed an optimized Class Activation Mapping (CAM) method known as LayerCAM~\cite{jiang2021layercam} to generate a heatmap for benign lesion images. LayerCAM is integrated following the final convolutional layer of DenseNet\cite{huang2017densely}, highlighting key features relevant to classification. This approach provides a rough approximation of the lesion area without the need for pixel-level labeling. The calculation of the prediction score for a target category ($c$) is described by Equation~\ref{eq:laycam1}, where $y^{c}$ is the score, $f^{c}(I,\theta)$ represents the classifier function with parameters $\theta$, and $I$ is the input image. Let $A$ denote the output feature map from the final convolutional layer, with $A^{k}$ being the $k$-th feature map within $A$. Each activation in $A^{k}$ at spatial position $(i,j)$ is represented by $A_{ij}^{k}$.
\begin{equation}
\begin{aligned}
y^{c}=f^{c}(I,\theta)
\end{aligned}
\label{eq:laycam1}
\end{equation}
To compute the gradient of the target category's prediction score with respect to a specific spatial location in $A^{k}$, we use Equation~\ref{eq:laycam2}:
\begin{equation}
\begin{aligned}
g_{ij}^{kc}=\frac{\partial y^{c}}{\partial A_{ij}^{k}}
\end{aligned}
\label{eq:laycam2}
\end{equation}
LayerCAM uniquely assigns weights to each spatial location in the feature map based on their importance. These weights are determined by the gradients, using positive gradients as weights and assigning zero to negative gradients (Equation~\ref{eq:laycam3}):
\begin{equation}
\begin{aligned}
w_{ij}^{kc}=ReLU(g_{ij}^{kc})
\end{aligned}
\label{eq:laycam3}
\end{equation}

The weighted activation for each position in the feature map is calculated using Equation~\ref{eq:laycam4}:
\begin{equation}
\begin{aligned}
\hat{A}_{ij}^{k}=w_{ij}^{kc}\cdot A_{ij}^{k}
\end{aligned}
\label{eq:laycam4}
\end{equation}

Finally, to obtain the class activation map, the adjusted activations $\hat{A}^{k}$ are linearly combined across the channel dimension and passed through a ReLU function (Equation~\ref{eq:laycam5}):
\begin{equation}
\begin{aligned}
M^{c}=ReLU(\sum_{k}\hat{A}^{k})
\end{aligned}
\label{eq:laycam5}
\end{equation}

%LayerCAM effectively utilizes deep learning to correlate key image features with potential disease areas, yielding valuable semantic information about lesions. Additionally, it can transform the class activation map into a binary image using a specific threshold. This thresholding process facilitates the initial identification of lesion location and size without requiring any extra annotations.

\subsection{Feature fusion and region synthesis} \label{sec:methods:fusion}

The morphology based algorithm is effective in contour extraction but faces challenges in medical image segmentation due to low contrast between lesions and surrounding tissues, complex lesion shapes and boundaries, and sensitivity to image noise. This sensitivity lead to incorrect clustering and erroneous segmentation.
Conversely, deep classification networks, through LayerCAM, can identify salient object regions but often suffer from imprecise activations. To address these limitations, we propose a fusion of traditional segmentation and deep learning methods, leveraging the strengths of both to construct a more accurate and complete lesion region, thereby enhancing segmentation accuracy.

\begin{figure}
  \centering
  \includegraphics[width=1\linewidth]{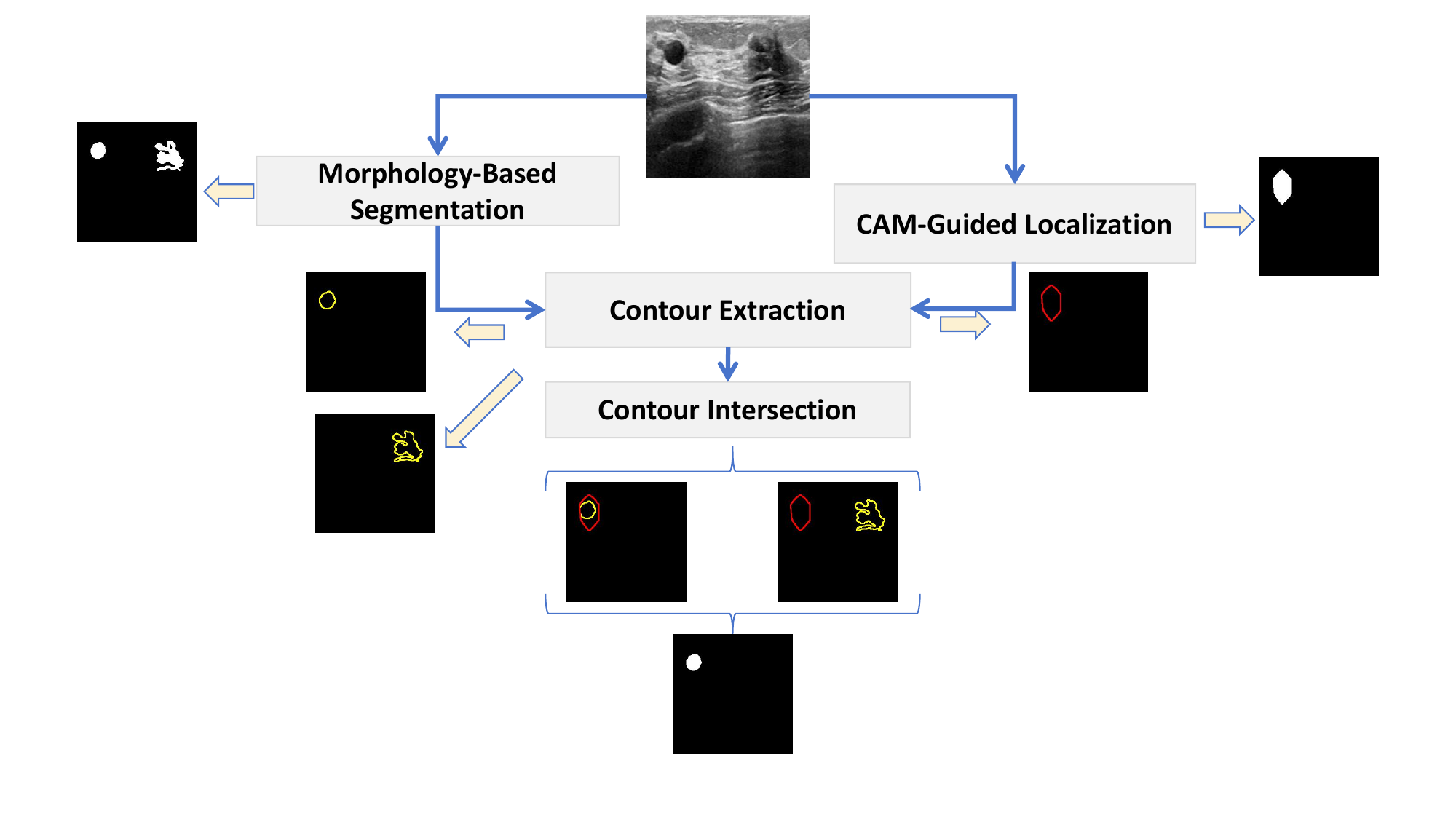}
  \caption{The feature fusion and lesion synthesis process. This step extracts the contours of each lesion in the traditional morphology-based segmentation results and determines the lesion with the largest area of intersection with the results of the CAM-guided lesion localization module as the synthesized result.}
  \label{fig:morphology_cam}
\end{figure}

Let $L^{M}=\{\ l_{1}^{M},...,l_{i}^{M},...,l_{n}^{M}\}$ represents the set of suspected lesions extracted via traditional morphology-based segmentation, and $L^{S}=\{\ l_{1}^{S},...,l_{j}^{S},...,l_{m}^{S}\}$ denote the set identified through CAM-guided lesion localization module. We extract the outline of each lesion in the $L^{M}$ and fuse them with lesions in the $L^{S}$ to identify the one with the maximum intersection area. This lesion is then selected as the final synthetic result for next step. 
If there's no intersection between $L^{M}$ and $L^{S}$, we consider $L^{S}$ as our segmentation outcome. This operation is mathematically expressed in Equation~\ref{eq:fuse}:
\begin{equation}
\begin{aligned}
p=\begin{cases}
L_{\max}^{M}, & if \quad L^{M}\cap L^{S}\neq0 \\
L^{S}, & otherwise \\
\end{cases}
\end{aligned}
\label{eq:fuse}
\end{equation}
where $L_{max}^{M}$ is the lesion with the largest overlap between $L^{M}$ and $L^{S}$. Figure~\ref{fig:morphology_cam} illustrates this fusion process. In this way, a collection of lesion regions that encompass both morphological and semantic information.

\subsection{SAM-Optimized lesion segmentation}\label{sec:methods:sam}
% Considering that the dataset we used is a training set for MedSAM and that the segmentation of primary brain lymphomas in the article\cite{guanghui2024comparing} showed that SAM segmentation performance outperforms the segmentation performance of MedSAM, we choose SAM to enhance the segmentation results. 
In this section, we detail the utilization of the SAM to enhance segmentation results obtained from synthesis regions. The initial segmentation, while incorporating morphological and CAM based semantic information, often lack precise lesion boundaries and areas. SAM, with its capability for high-precision segmentation, is proposed as a powerful tool to refine the segmentation result. SAM's architecture comprises three key components: an image encoder, a prompt encoder, and a mask decoder. The image encoder uses a scalable Vision Transformer (ViT) pre-trained by MAE~\cite{he2022masked}, adept at processing high-resolution inputs. Its primary function is to transform the target image into a feature space representation. For BUS image segmentation, we have devised two strategies to generate box and point prompts to integrate with SAM. These prompts are then fed into the encoder.
The mask decoder plays a crucial role in integrating the embeddings produced by both the image and prompt encoders. It decodes the final segmentation mask from the combined feature map of these embeddings. This process effectively aligns the image embedding, prompt embedding, and output token to generate a detailed and accurate mask, thereby enhancing the overall quality and precision of the segmentation.

In our research, the BUS images often contain complex biological structures and are susceptible to various noise sources. Direct application of the SAM proved insufficient for medical image segmentation due to these complexities. However, we discovered that using intermediate results as seed signals in SAM significantly improves its efficacy.
We used the original BUS image as input to SAM and experimented with two ways to interact with SAM: box prompt and point prompt. In box prompt segmentation, for each breast ultrasound (BUS) image, we generate the smallest enclosing rectangle from the fused pseudo-label information. This rectangular data is then fed as a seed region signal into SAM's prompt encoder, where it is transformed into embedding vectors. A mask decoder, leveraging these embeddings, segments the lesion mask from the BUS image.
The second method, point prompt segmentation, involves generating random points on the fused pseudo-labels. The coordinates of these points are input into SAM's prompt encoder, leading to enhanced pseudo-labels post-SAM segmentation.%后处理方法需要详细说明
Moreover, we observed that the lesion areas segmented by SAM often contain holes. To rectify this, we apply a morphological reconstruction post-processing operation to refine the topological error. This method involves iterative expansion and erosion operations. In the post-processing stage, we performed an iterative expansion operation on the SAM-optimized segmented lesion regions to identify and fill these gaps, resulting in more accurate lesion segmentation outcomes.

\section{Experiments and results} \label{sec:experiments}

\subsection{Dataset}
The Breast Ultrasound Image (BUSI) dataset~\cite{al2020dataset} is a classification and segmentation resource comprising ultrasound images from 600 female patients aged 25 to 75, collected in 2018. It contains 780 PNG images, categorized into normal, benign, and malignant classes, with 133 normal, 437 benign, and 210 malignant images.
Our research is centered on understanding the early mechanisms of breast health and disease, primarily aimed at early detection and prevention of breast cancer. The segmentation of benign and normal images is crucial in identifying potential early abnormalities, thereby enhancing both prevention and early diagnosis for patients. Due to incomplete lesion labeling in the dataset, such as instances of multiple lesions with only one marked, we undertook a secondary selection process. This refined the dataset to 123 normal and 365 benign cases. To counter the imbalanced data distribution, we augmented the normal images using techniques like flipping and rotation. The dataset was then randomly divided into a training set with 390 images (around 80\%) and a testing set with 98 images (around 20\%). %All experimental images were resized to $256 \times 256$ dimensions for consistency in analysis.

To further validate our generalization, we used a second breast ultrasound dataset, referred to as Dataset B, proposed by Yap et al.~\cite{yap2020breast}. This dataset contains 163 images with an average size of 760×570 pixels, collected using the Siemens ACUSON Sequoia C512 system. It includes 109 benign cases and 54 malignant cases. 

\subsection{Implementation details}
For these two experimental datasets, all the images were resized to 256×256 pixels. Note that all ablation experiments and parameter selections were performed using the BUSI dataset~\cite{al2020dataset}, while Dataset B was used solely for additional evaluation.

Our implementation utilizes the PyTorch framework~\cite{paszke2019pytorch} and is trained on a single NVIDIA GeForce RTX 3090 (24GB). The morphology-based segmentation module employs K-means clustering with $k=2$ for lesion extraction, using a binarization threshold of 90. The classification model we used is DenseNet121~\cite{huang2017densely}, pre-trained on ImageNet~\cite{deng2009imagenet}. We utilize Stochastic Gradient Descent (SGD) as the optimizer, with a weight decay of 0.0004 and momentum of 0.9. The learning rate starts at $1e-4$, and the model is trained over 100 epochs with a minimum batch size of 16. We employ several evaluation metrics: Dice score, the 95th percentile of Hausdorff Distance (HD95), and Intersection over Union (IoU). For each of these metrics, we calculate and report both the mean and the 95\% confidence interval. The confidence intervals are determined using bootstrap analysis, involving 5000 resampling iterations to ensure statistical robustness and accuracy in our results.

Our ablation experiments, detailed in Section~\ref{sec:experiment:results}, were meticulously designed to assess the performance of each module. We summarize the experimental setup and final setting as follows. We employ LayerCAM to determine the approximate location of the lesion and establish a binarization threshold at 200. In the final experiments, we incorporate the ViT-H model version of SAM. 
Our complete codes are public available for download at: \url{https://github.com/YueXin18/MorSeg-CAM-SAM}.

\subsection{Experimental results} \label{sec:experiment:results}

\subsubsection{CAM based lesion location}
To refine lesion localization and segmentation, we evaluated five class activation mapping (CAM) methods: Grad-CAM~\cite{selvaraju2017grad}, AblationCAM~\cite{ramaswamy2020ablation}, LayerCAM~\cite{jiang2021layercam}, EigenGradCAM~\cite{muhammad2020eigen}, and Grad-CAM++~\cite{chattopadhay2018grad}. 
% We determined a threshold of 200 by calculating the lesion intersection ratio of the CAM methods results to the results of the morphology-based segmentation. The grayscale maps obtained from the CAM are transformed into binary segmentation maps using this threshold. 
We established a threshold by calculating the lesion intersection ratio from the results of the CAM methods compared to those from morphology-based segmentation. Using this threshold, the grayscale maps obtained from the CAM are converted into binary segmentation maps. In our experiments, the threshold was computationally determined to be set at 200.
%CAM二值化问题：计算与形态学部分结果的交并比确定阈值
%In the threshold selection section, we developed a selection strategy. We calculated the intersection ratio of lesions for the binary maps of the five CAM methods obtained at four thresholds from 180 to 210 with the resultant maps obtained in the morphology-based segmentation,  detailed in Section~\ref{sec:methods:morphology}. The highest intersection ratio between the LayerCAM localization results and the morphology-based segmentation results was obtained when the threshold was 200. Informed by these results, we have chosen LayerCAM, set at a threshold value of 200, as the module for advancing to the subsequent experiment.

We also validated the threshold selection method and results using ground truth. We conducted comparisons across five CAM related methods at different thresholds from 180 to 210. However, due to Grad-CAM's inability to segment lesions at a threshold of 210, we limited the comparison at this threshold to the other four methods. The experimental results validating the threshold selection on the BUSI dataset are detailed in Table~\ref{tab:result_cam}, with the best outcomes highlighted in bold. 
%解释LayerCAM阈值的影响。例如，在表 1 中，为什么阈值等于 200 时，LayerCAM 在 HD95 上的结果不如阈值等于 210 时的结果好，而 Dice 和 IoU 的结果却更好？ 
It can be seen that among the five CAM methods, the initial localization of lesions using LayerCAM has the best performance. When the threshold is 200, the results of LayerCAM on HD95 are not as good as when the threshold is equal to 210, but the results of Dice score and IoU metrics are better. This is because the threshold has an effect on the selection of significant activations, and more activations may be included when the threshold is set to 200. Whereas HD95 focuses on the difference between the boundary predicted by the model and the true boundary. Therefore, more activations may lead to larger boundary errors, which may affect HD95. Dice score and IoU metrics focus more on the overall picture of the segmentation results than HD95, and these additional activations may also provide more information that may be beneficial for metrics such as Dice score and IoU.

\begin{figure}
\centering
\includegraphics[width=1\linewidth]{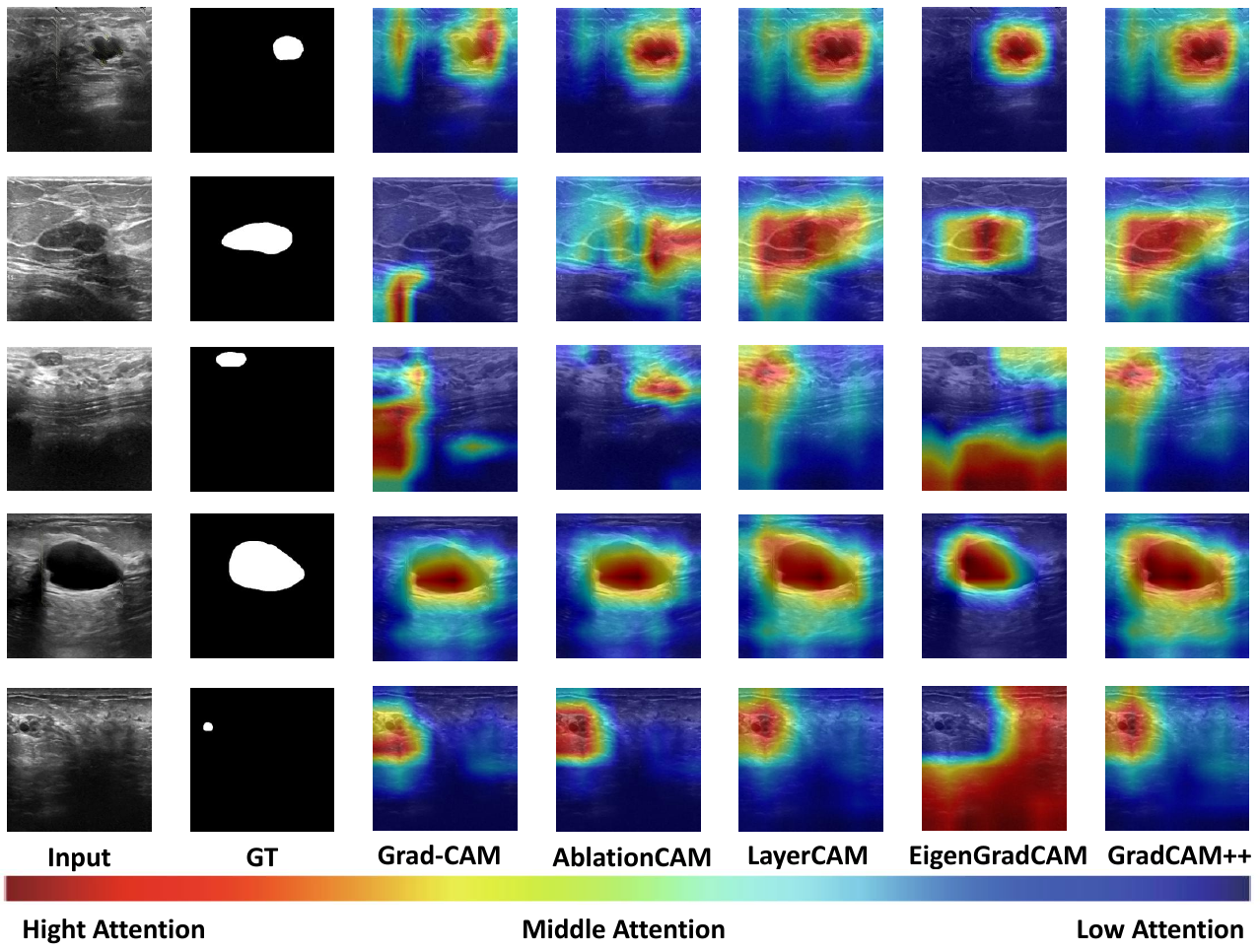}
\caption{Visualization of different Class Activation Mapping (CAM) Methods at a threshold value of 200 on the BUSI dataset.}
\label{fig:cam_result}
\end{figure}

LayerCAM's superiority lies in its ability to assign distinct weights to each spatial location, thereby acknowledging the varying significance of the class of interest. This feature enables LayerCAM to eliminate background noise and retain reliable object localization information. The visualization of the results from various CAM models on the BUSI dataset is presented in Figure~\ref{fig:cam_result}. 

\begin{table*}[hbtp!]
\begin{center}
\caption{Performance comparison of CAM methods under different thresholds on the BUSI dataset. The models that achieved the best performance are highlighted in bold.}
\begin{tabular}{|l|l|l|l|l|}
\hline
 Threshold &Methods & Dice($\%$)  &HD95 &IoU($\%$)\\
\hline
 \multirow{5}{*}{180}& Grad-CAM & 35.60 [30.58,40.58]
& 50.65 [42.78,59.96]
&23.86 [20.10,27.71]
\\
 & AblationCAM  & 36.69 [31.61,41.70]
& 49.37 [41.88,58.37]
&24.89 [20.95,28.92]
\\
 & LayerCAM& 36.32 [31.15,41.78]
& 56.78 [48.01,66.74]
&24.89 [20.67,29.39]
\\
 & EigenGradCAM  & 36.91 [31.16,42.71]
& 62.86 [48.88,78.41]
&25.73 [21.33,30.31]
\\
 & Grad-CAM++  & 35.70 [30.69,40.98]
& 54.13 [46.50,62.49]
&24.24 [20.19,28.60]
\\
\hline
 \multirow{5}{*}{190}& Grad-CAM & 35.98 [31.09,40.73]
& 49.20 [41.37,58.41]
&23.99 [20.33,27.70]
\\
 & AblationCAM  & 37.39 [32.46,42.28]
& 47.67 [40.33,56.32]
&25.27 [21.50,29.14]
\\
 & LayerCAM& 37.86 [32.72,43.17]
& 52.43 [44.01,61.88]
&25.93 [21.79,30.27]
\\
 & EigenGradCAM  & 37.81 [32.00,43.56]
& 61.87 [47.77,77.32]
&26.41 [22.00,30.98]
\\
 & Grad-CAM++  & 37.24 [32.20,42.46]
& 50.77 [43.24,59.22]
&25.35 [21.32,29.59]
\\
\hline
 \multirow{5}{*}{200}&Grad-CAM & 35.78 [30.93,40.53]
&47.05 [39.20,56.19]
&23.81 [20.21,27.45]
\\
 &AblationCAM  & 37.72 [32.85,42.57]
&46.32 [38.87,54.92]
&25.41 [21.72,29.22]
\\
 &\textbf{LayerCAM}  & \textbf{40.40 [35.28,45.57]}&45.60 [38.17,53.88]
&\textbf{27.78 [23.66,32.01]}\\
 & EigenGradCAM  & 38.35 [32.48,44.22]
&61.18 [46.94,76.70]
&26.86 [22.33,31.48]
\\
 &Grad-CAM++  & 38.78 [33.69,43.82]
&47.36 [39.97,55.59]
&26.46 [22.48,30.56]
\\
\hline
 \multirow{4}{*}{210}&AblationCAM  & 36.86 [31.85,41.70]
&45.15 [37.52,53.91]
&24.75 [20.97,28.54]
\\
 &LayerCAM& 40.34 [35.29,45.28]
&\textbf{45.09 [37.50,53.77]}&27.61 [23.73,31.48]
\\
 & EigenGradCAM  & 37.65 [31.86,43.53]
&60.78 [46.35,76.39]
&26.25 [21.74,30.81]
\\
 &Grad-CAM++  & 39.50 [34.44,44.53]
&46.16 [38.25,55.19]
&26.99 [23.05,31.01]
\\
 \hline
\end{tabular}
\label{tab:result_cam}
\end{center}
\end{table*}

\subsubsection{Ablation experiments} \label{subsubsec:ablation}

In our ablation study, we systematically evaluated the performance impact of various components and configurations within our proposed framework, and presented the results on the BUSI dataset. This approach helps in understanding the contribution of each module to the overall segmentation task. 
We experiment with two prompting methods for SAM: the point prompt and the box prompt. In the point prompt approach, we randomly generate 10 points within the synthesized area described in Section~\ref{sec:methods:fusion} and input these into the SAM model. For the box prompt method, we construct the smallest enclosing rectangle around the synthesized lesion as the box prompt. And then subsequently fed into SAM for segmentation. 

The results of the ablation experiments are shown in Table~\ref{tab:result_ablation}, with bold indicating the best results. Each configuration was meticulously assessed to determine its contribution to the effectiveness of the segmentation task. This structured approach not only underscores the individual significance of the modules but also illustrates their combined impact in our comprehensive framework. 

\begin{figure*}
  \centering
  \includegraphics[width=1\linewidth]{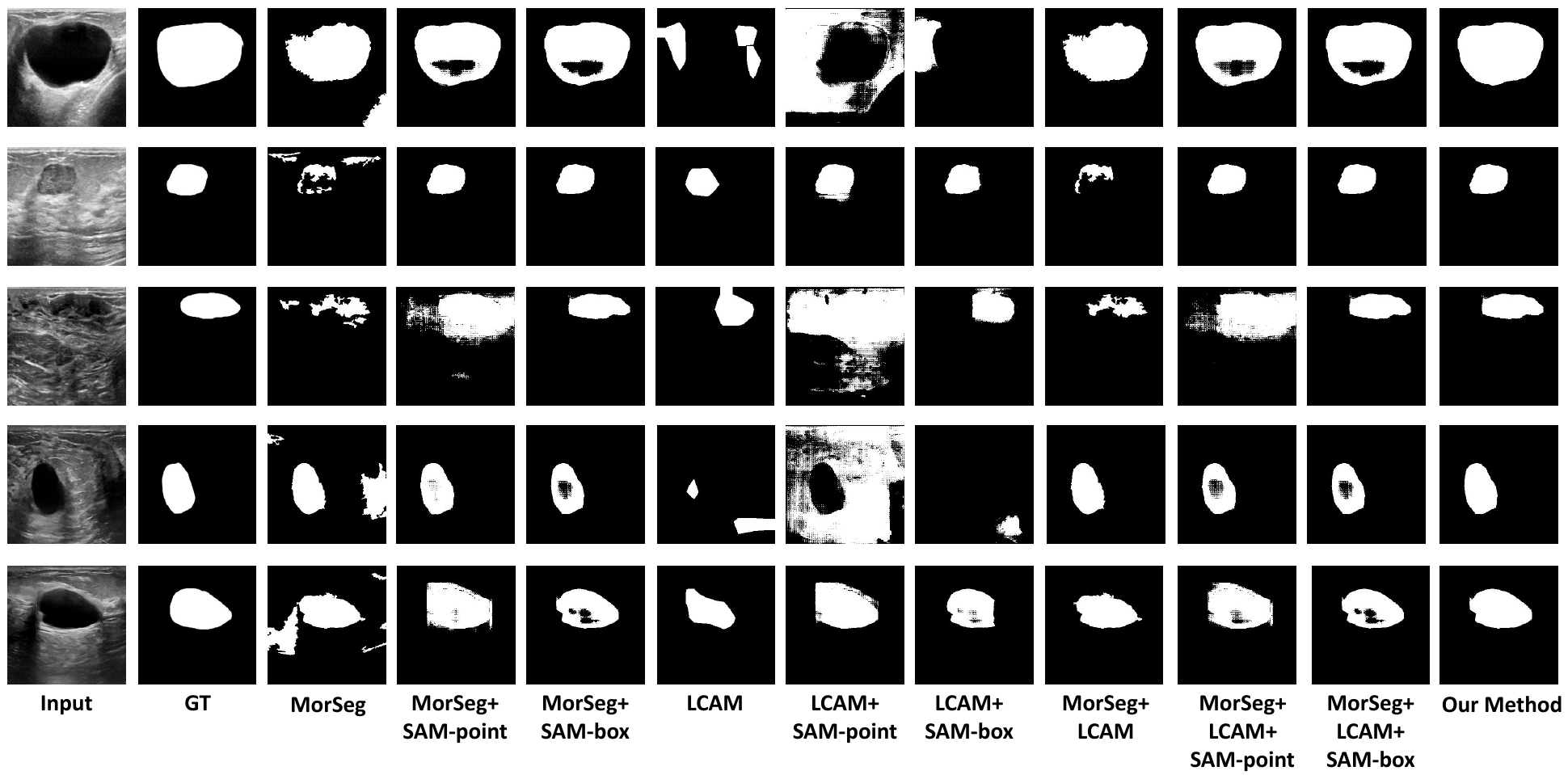}
  \caption{Visualization results from ablation study on the BUSI dataset.}
  \label{fig:result_ablations}
\end{figure*}

The results presented in the table indicate that using a box prompt as input for SAM yields better performance compared to a point-based prompt. Furthermore, the most effective segmentation results for this task are attained by integrating all these modules.
The visualization of the ablation experiment's results is presented in Figure~\ref{fig:result_ablations}. It is evident from these visualizations that each of our proposed modules contributes to enhancing the performance of lesion segmentation.
A notable observation from Figure~\ref{fig:result_ablations} is the tendency of SAM-segmented results to exhibit holes in certain areas. This issue may stem from the manner in which prompts are provided to the model, underscoring the necessity for final post-processing steps to address these gaps. The performance of our model shown in Table~\ref{tab:result_vs_supervised} is after post-processing to fill the holes inside the segment area.

\begin{table*}
\centering
\caption{Ablation results of segmented breast lesions on the BUSI dataset. MorSeg denotes morphological-based segmentation, while LCAM denotes the layerCAM module. SAM-p and SAM-b refer to the SAM module based on point and box prompts, respectively.}
\label{tab:result_ablation}
\begin{tabular}{|l|l|l|l|c|c|c|} 
\hline
\multicolumn{4}{|c|}{Modules}                                                                                                                                 & \multirow{2}{*}{Dice(\%)} & \multirow{2}{*}{HD95}  & \multirow{2}{*}{IoU(\%)}  \\ 
\cline{1-4}
MorSeg                                & LCAM                                  & SAM-p                                 & SAM-b                                 &                           &                        &                           \\ 
\hline
{\checkmark} &                                       &                                       &                                       & 45.15 [39.23,51.02]       & 104.70 [93.72,115.36]  & 32.95 [27.91,38.28]       \\
{\checkmark} &                                       & {\checkmark} &                                       & 45.66 [36.65,55.05]       & 82.98 [67.38,98.48]    & 39.29 [30.80,48.05]       \\
{\checkmark} &                                       &                                       & {\checkmark} & 45.49 [35.74,55.24]       & 64.45 [51.17,78.14]    & 40.25 [31.27,49.27]       \\
                                      & {\checkmark} &                                       &                                       & 40.40 [35.28,45.57]       & 45.60 [38.17,53.88]    & 27.78 [23.66,32.01]       \\
                                      & {\checkmark} & {\checkmark} &                                       & 22.21 [16.12,28.97]       & 124.76 [112.39,136.40] & 16.55 [11.10,22.69]       \\
                                      & {\checkmark} &                                       & {\checkmark} & 58.22 [50.54,65.80]       & 35.74 [27.22,45.12]    & 48.57 [41.19,55.91]       \\
{\checkmark} & {\checkmark} & &                                       & 67.04 [60.31,73.20]       & 29.33 [21.79,37.75]    & 56.14 [49.77,62.21]       \\
{\checkmark} & {\checkmark} & {\checkmark} &                                       & 70.45 [63.04,77.24]       & 36.30 [25.58,48.14]    & 61.73 [54.79,68.69]       \\
{\checkmark} & {\checkmark} &                                       & {\checkmark} & \textbf{74.10 [66.84,80.68]}       & \textbf{25.63 [18.13,34.25]}    & \textbf{65.82 [58.71,72.27]}       \\
\hline
\end{tabular}
\end{table*}

\subsubsection{Comparison of SAM in different versions with different prompts}
\begin{figure}
  \centering
  \includegraphics[width=1\linewidth]{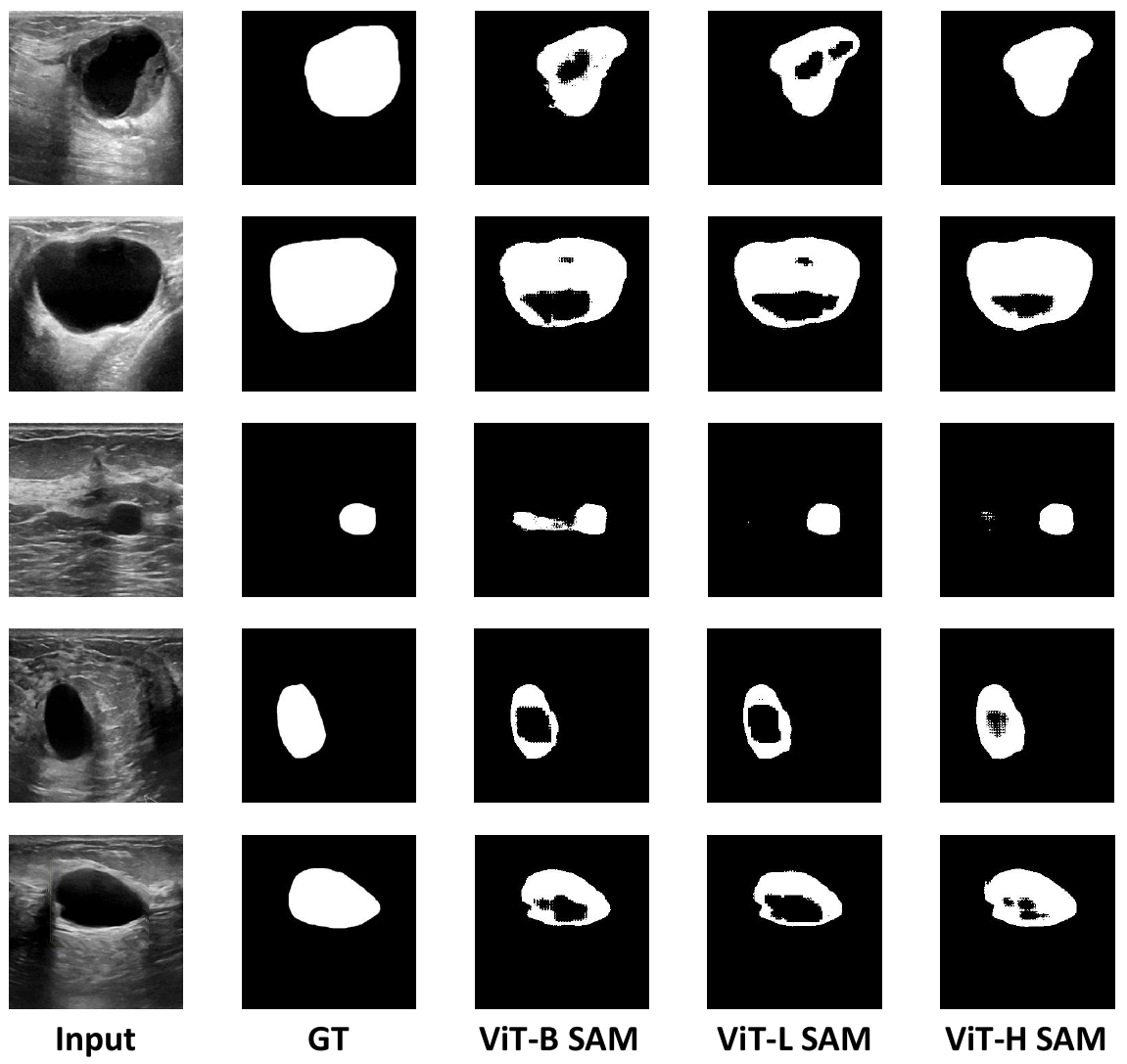}
  \caption{Visualization results of different versions of SAM segmentation  on the BUSI dataset.}
  \label{fig:result_sam}
\end{figure}

In our study, we evaluated three models from the SAM series, each with different parameter counts: ViT-B (91M parameters), ViT-L (308M), and ViT-H (636M). 
In this experiment, we utilize the box prompt for the SAM model, as the results presented in Table~\ref{tab:result_ablation} indicate superior performance of the box prompt compared to the point prompt. Comparative experiments to evaluate their performance were conducted, with the results detailed in Table~\ref{tab:result_sam}, where the best results are highlighted in bold. 
The performance of these results is also illustrated in Figure~\ref{fig:result_sam}.
The experimental findings suggest a trade-off between resource consumption and performance efficiency, indicating that the choice of model should align with the capabilities of the available experimental equipment. Based on these considerations, we selected ViT-H SAM for our final segmentation tasks.
The observations from Figure~\ref{fig:result_sam} highlight that holes are present in all versions of SAM predictions, underscoring the importance of post-processing when utilizing SAM for accurate results.

\begin{table*}[hbtp!]
\begin{center}
\caption{Comparison of segmentation performance of different versions of the SAMs based on box prompt on the BUSI dataset. AVG times indicates the average time to process a BUS image.}
\begin{tabular}{|l|l|l|l|l|}
\hline
Methods & Dice($\%$)  &HD95 &IoU($\%$)& AVG Time(s)\\
\hline
ViT-B SAM& 71.04 [63.76,77.63]  &26.66 [19.03,35.36]&62.10 [54.93,68.65]& 0.1470\\
ViT-L SAM& 73.36 [66.24,79.84]  &\textbf{24.70 [17.45,33.10]}&64.52 [57.76,70.75]& 0.2934\\
ViT-H SAM& \textbf {74.10 [66.84,80.68]}  &25.63 [18.13,34.25]&\textbf{65.82 [58.71,72.27]}& 0.4817\\
\hline
\end{tabular}
\label{tab:result_sam}
\end{center}
\end{table*}
\subsubsection{Compare with other deep learning models}

In our study, we evaluated the performance of our proposed framework by comparing it with three methods on two publicly available datasets.  The methods used for comparison are UNet~\cite{ronneberger2015u}, Deeplabv3+~\cite{chen2018encoder}, and AffinityNet~\cite{ahn2018learning}. UNet and Deeplabv3+ are fully supervised networks, whereas AffinityNet operates under a weakly supervised paradigm with only image-level labeling. For an equitable comparison, all networks were retrained on the BUSI dataset and Dataset B. Quantitative comparison results are presented in Table~\ref{tab:result_vs_supervised}. 
The experimental results reveal that the two supervised learning methods (UNet and Deeplabv3+) exhibit comparable performance in both Dice score across the BUSI and Dataset B datasets, as indicated by the overlap of their 95\% confidence intervals.
Compared to these supervised learning methods, our model demonstrates slightly lower performance, with a Dice score that is 4.01\% points lower on the BUSI dataset and 8.67\% points lower on Dataset B. Despite these differences, the overlapping confidence intervals suggest that these variations are not statistically significant~\cite{jurdi2023confidence, el2023precise}.
In terms of HD95, our model surpasses UNet by 2.61 on the BUSI dataset and shows a reduction of 7.95 compared to Deeplabv3+, indicating superior precision in delineating lesion boundaries. On Dataset B, the HD95 of our model is similar to those of the supervised learning models, being only 3.75 higher than Deeplabv3+. 

Compared to the weakly supervised method, AffinityNet, our method demonstrates significant improvements across all evaluation metrics. The inferior performance of AffinityNet underscores both the complexity of this task and the advantage of our method's comprehensive strategy for utilizing information, which includes both lesion contours and semantic details from breast images. These outcomes collectively underscore the effectiveness and precision of our proposed framework in medical image segmentation tasks. 

\begin{table*}[hbtp!]
\begin{center}
\caption{Performance comparison with some breast lesion segmentation methods on two public datasets. SL means supervised learning and WSL means weakly supervised learning. }
\begin{tabular}{|l|l|l|l|l|l|} 
\hline
Datasets     &Models     & Training  & Dice($\%$)                        &HD95 &IoU($\%$)\\ 
\hline
\multirow{4}{*}{BUSI}     &UNet       & \multirow{2}{*}{SL}        & 78.31 [71.77,84.28]&\textbf{21.66 [11.05,35.54]}&\textbf{70.51 [63.61,76.77]}\\ 
    &Deeplabv3+  &                            & \textbf{78.40 [72.78,83.21]}&32.22 [17.79,49.91]&68.49 [63.10,73.23]\\ \cline{2-6}

     &AffinityNet & \multirow{2}{*}{WSL} & 16.14 [10.98,21.73]&78.15 [68.08,89.41]&10.98 [7.03,15.35]\\
     &Proposed model  &                            & 74.39 [67.09,81.02]&24.27 [16.67,32.85]&66.27 [59.09,72.91]\\
\hline
\multirow{4}{*}{Dataset B}&UNet       & \multirow{2}{*}{SL}        & \textbf{82.63[76.86,87.86]
}&21.73[10.65,35.36]
&\textbf{72.38[65.01,79.41]
}\\ 
    &Deeplabv3+  &                            & 78.91[70.26,86.39]
&\textbf{18.72[6.69,34.28]}&68.90[59.06,77.82]
\\ \cline{2-6}

     &AffinityNet & \multirow{2}{*}{WSL} & 32.92[20.44,45.47]
&112.73[87.55,137.34]
&23.88[14.57,33.32]
\\
     &Proposed model  &                            & 73.96[60.65,85.06]
&22.47[9.70,39.05]
&65.76[53.14,76.87]
\\
\hline
\end{tabular}
\label{tab:result_vs_supervised}
\end{center}
\end{table*}

\begin{figure}
  \centering
  \includegraphics[width=1\linewidth]{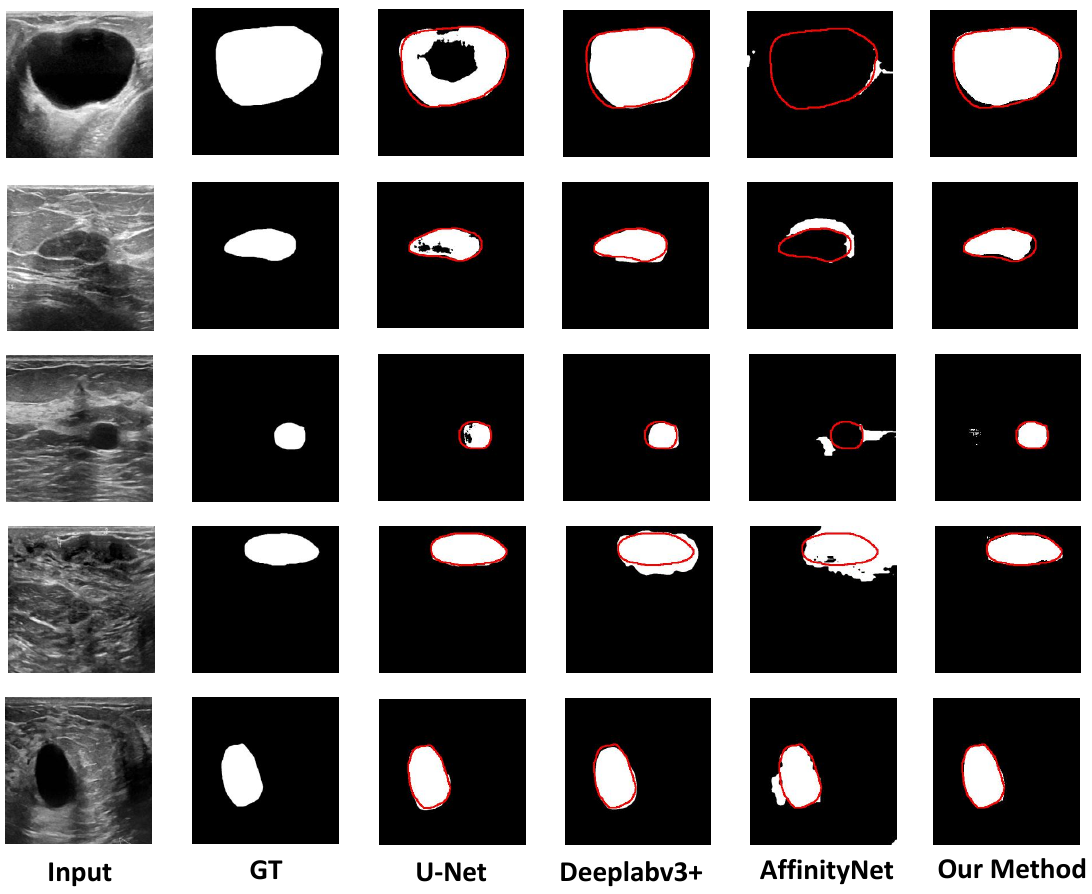}
  \caption{Comparative experiment results visualization on the BUSI dataset. Red contour lines depict the lesion edges as delineated in the ground truth labels.}
  \label{fig:result_methods}
\end{figure}

Figure~\ref{fig:result_methods} offers a visual comparison of segmentation results on the BUSI dataset from our method against UNet, Deeplabv3+, and AffinityNet, using representative cases from the test set. This comparison distinctly highlights the sensitivity of our method to lesion contours. As observed in Figure~\ref{fig:result_methods}, our proposed framework excels in segmenting contours and smaller lesions, outperforming both UNet and Deeplabv3+ in these aspects. The figure further reveals that the fully supervised methods, UNet and Deeplabv3+, tend to be influenced by background noise during segmentation, leading to the extraction of some non-lesion tissues. Our method effectively addresses this issue by removing incorrect segmentations through our refined process of filtering suspected lesions. In comparison with AffinityNet, which operates under a similar weakly supervised framework, our method demonstrates superior accuracy in producing segmentation masks. This indicates the effective extraction and utilization of semantic features from breast images in our approach. Overall, our proposed method achieves competitive results, particularly notable for its lower reliance on extensive annotation, compared to traditional fully supervised segmentation techniques.

\section{Discussion}

Our model comprises four different modules: morphological feature segmentation, CAM-guided localization, feature fusion, and final SAM-optimized segmentation. Optimal results necessitate coordinated functioning between these modules, which presented a challenge in our study. Nonetheless, our objective is to maximize the utilization of medical prior knowledge to minimize the reliance on detailed pixel-level annotation. The ablation experiment validates the essential role of each module and provides insights into their individual performances, such as using only morphology and CAM, or the efficiency of a lightweight SAM. This flexibility enables researchers to tailor module combinations to suit specific needs and contexts.
In the CAM-guided localization module, LayerCAM effectively utilizes deep learning to correlate key image features with potential disease areas, yielding valuable semantic information about lesions. Additionally, it can transform the class activation map into a binary image using a specific threshold. This thresholding process facilitates the initial identification of lesion location and size without requiring any extra annotations.
%请简要谈谈这种方法的运行效率。与其他模型相比，它在效率方面有哪些优势？
In terms of operational efficiency, our proposed segmentation framework is typically faster to train due to the simpler and lighter-weight model structure design using only category-level labels without the need to deal with detailed pixel-level annotations as in end-to-end supervised learning.

This study aims to develop an effective primary screening method for breast cancer identification, potentially reducing misdiagnosis and overtreatment, particularly in resource-constrained environments. Our focus was solely on benign tumors in mammography for experimental data. Although the performance of our proposed method on the BUSI dataset differs by only 4\% in Dice score compared to supervised learning, the considerable overlap in the performance confidence intervals of both models suggests that the difference might not be statistically significant~\cite{jurdi2023confidence, el2023precise}. The promising performance of our approach paves the way for incorporating a broader spectrum of data types and diseases in future research.

\section{Conclusion} \label{sec:conclusion}
In this study, we introduce a novel, morphology-enhanced CAM-guided SAM framework for weakly supervised segmentation of breast lesions from ultrasound images. Our methodology, evaluated using the BUSI  and Dataset B public dataset, effectively segments lesions with image-level labeling. The framework capitalizes on a priori knowledge of breast lesion morphology for contour extraction and incorporates semantic feature extraction and lesion localization using a CAM-based approach. We explored various class activation mapping techniques, ultimately integrating LayerCAM for highlighting lesion regions.
Leveraging the strengths of both segmentation methods, we fuse the extracted information for more accurate and smoother segmentation. The SAM model serves as a powerful segmentation enhancement tool, refining these synthesis results. A final post-processing step is applied for further enhancement.
Our approach demonstrates notable effectiveness, achieving a Dice score of 74.39\%, and a 95th percentile Hausdorff Distance (HD95) of 24.27 on the BUSI dataset. These results not only affirm the validity and superiority of our method but also show its competitive edge over fully supervised network like Deeplabv3+ in boundary segmentation accuracy, while significantly outperforming weakly supervised networks that rely solely on image-level labels.
In the future research, we aim to expand this framework's application to lesion segmentation in other medical imaging datasets, further advancing the field of medical image analysis.

% \section{Dataset and Code Availability}
% \begin{itemize}
%     \item The dataset employed in our experiments is publicly accessible via: \url{https://scholar.cu.edu.eg/?q=afahmy/pages/dataset}
%     \item The SAM models used in our research are also publicly available at: \url{https://github.com/facebookresearch/segment-anything}. 
%     \item Additionally, our complete code, encompassing both the model architecture and checkpoints, is available for download at: \url{https://github.com/YueXin18/MorSeg-CAM-SAM}.
% \end{itemize}

\section*{Acknowledgment}\label{sec:acknowledgments}
This study is supported by a research project from the Beijing Postdoctoral Research Foundation (No. 2022zz075). Guanghui Fu is supported by a Chinese Government Scholarship provided by the China Scholarship Council (CSC).

%\bibliography{references.bib}
%\bibliographystyle{IEEEtran}
\bibliography{references} 
\bibliographystyle{spiebib}

\end{document}